\documentclass[sigconf]{acmart}

\usepackage{booktabs}
\usepackage{multirow}
\usepackage{pifont}
\usepackage{siunitx}

\makeatletter
\def\@projectpage{}
\newcommand{\projectpage}[1]{\gdef\@projectpage{#1}}
\usepackage{etoolbox}
\pretocmd{\@printtopmatter}{%
  \ifx\@projectpage\@empty\else
    \global\setbox\mktitle@bx=\vbox{%
      \unvbox\mktitle@bx
      \vskip 6\p@
      \centerline{\Large \url{\@projectpage}}%
      \vskip 12\p@
    }%
  \fi
}{}{}
\makeatother

\AtBeginDocument{%
  }

\renewcommand\footnotetextcopyrightpermission[1]{}

\setcopyright{none}
\acmISBN{}
\acmDOI{}

\begin{document}

\title{Toward Robust and 3D-Aware RGB-NIR Imaging in the Dark}

\author{Muyao Niu}
\affiliation{%
  \institution{The University of Tokyo}
  \city{Tokyo}
  \country{Japan}
}
\email{muyao.niu@gmail.com}

\author{Mingze Ma}
\affiliation{%
  \institution{Adelaide University}
  \city{Adelaide}
  \country{Australia}
}
\email{mingze.ma@adelaide.edu.au}

\author{Yifan Zhan}
\affiliation{%
  \institution{The University of Tokyo}
  \city{Tokyo}
  \country{Japan}
}
\email{zhan-yifan@g.ecc.u-tokyo.ac.jp}

\author{Qingtian Zhu}
\affiliation{%
  \institution{The University of Tokyo}
  \city{Tokyo}
  \country{Japan}
}
\email{qtzhu@g.ecc.u-tokyo.ac.jp}

\author{Zhihang Zhong}
\affiliation{%
  \department{School of Artificial Intelligence (SAI)}
  \institution{Shanghai Jiao Tong University}
  \city{Shanghai}
  \country{China}
}
\email{zhongzhihang@sjtu.edu.cn}

\author{Wei Guo}
\affiliation{%
  \institution{The University of Tokyo}
  \city{Tokyo}
  \country{Japan}
}
\email{guowei@g.ecc.u-tokyo.ac.jp}

\author{Chang Wen Chen}
\affiliation{%
  \institution{Hong Kong Polytechnic University}
  \city{Hong Kong SAR}
  \country{China}
}
\email{changwen.chen@polyu.edu.hk}

\author{Yinqiang Zheng}
\correspondingauthor
\affiliation{%
  \institution{The University of Tokyo}
  \city{Tokyo}
  \country{Japan}
}
\email{yqzheng@ai.u-tokyo.ac.jp}

\projectpage{https://github.com/MyNiuuu/3DarkFusion}

\renewcommand{\shortauthors}{Muyao Niu et al.}

\begin{abstract}
  Robust low-light imaging remains challenging for the community. Recent studies have explored fusing Near-Infrared (NIR) with noisy RGB to achieve improved enhancement, yet most methods depend on carefully curated training data pairs, with limited robustness under different scenarios. This paper offers a new perspective for RGB-NIR low-light imaging by incorporating 3D-aware neural modeling. Without using clean RGB supervision, a powerful model can be optimized to implicitly fuse extremely noisy RGB observations with NIR cues in 3D space, effectively recovering clean RGB images. The proposed model obviates the requirement for clean RGB data collection, generalizes across different noise levels. Extensive evaluations on synthetic and real data demonstrate its superiority. Codes available: \url{https://github.com/MyNiuuu/3DarkFusion}
\end{abstract}

\maketitle

\section{Introduction}
\label{sec:intro}

Robust dark imaging remains a fundamental challenge in computer vision. Despite rapid advances in camera hardware, noise interference under extreme low-light conditions continues to degrade image quality. Conventional strategies, such as long-exposure and burst photography, are prone to motion blur~\cite{niu2026motion} and visibility issues~\cite{xiong2021seeing,niu2023visibility}, limiting their practicality.

To mitigate these issues, numerous low-light enhancement methods have been proposed, ranging from traditional denoising operators~\cite{buades2005non,dabov2007image,dong2012nonlocally,jia2019comdefend,xu2018trilateral,dong2012nonlocal,gu2014weighted} to modern deep learning–based networks~\cite{zamir2021multi,anwar2019real,chen2022simple,wang2022uformer,zamir2022restormer}.
Among these, several image- and video-based RGB–NIR low-light enhancement methods have recently attracted increasing attention~\cite{jin2022darkvisionnet,sheng2023structure,wang2024rffnet,xu2024nir} due to the invisibility of near-infrared signals and their complementary structural cues in darkness. However, existing approaches still face critical limitations. Supervised models are typically trained on ``noisy RGB–NIR–clean RGB'' triplets from specific domains, resulting in poor cross-domain generalization and limited robustness under varying noise levels. This leads to a natural question: Can robust low-light enhancement be achieved using only NIR and noisy RGB observations, without relying on any form of clean RGB supervision?

\begin{figure*}
\centering
\setlength{\abovecaptionskip}{2mm}
  \includegraphics[width=\textwidth]{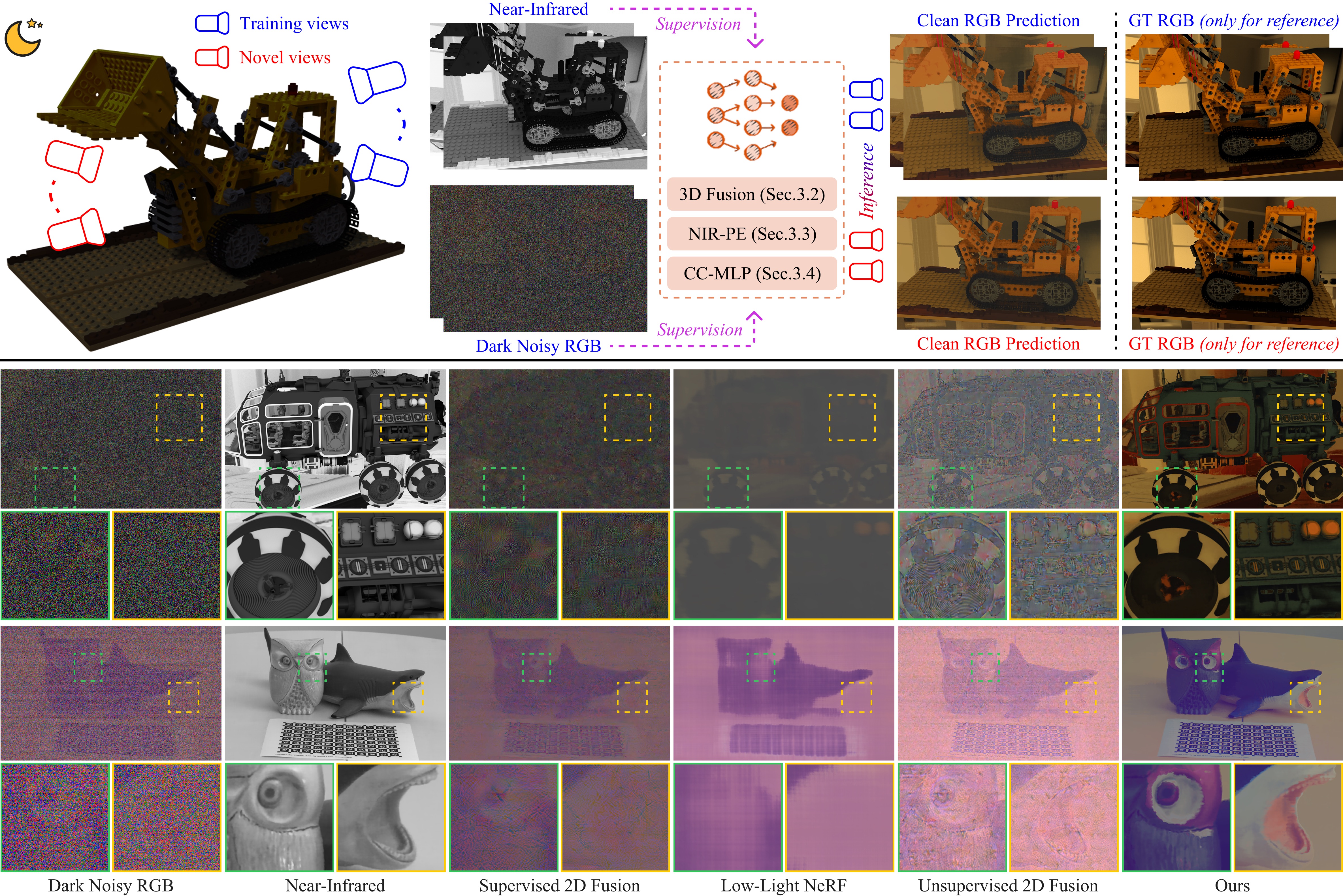}
  \caption{ 
  \textbf{Top:} Given extremely noisy multi-view RGB observations in darkness, the proposed method recovers the clean RGB scene using Near-Infrared. NO clean RGB is used for optimizing the model. \textbf{Bottom:} Visual comparisons against methods from various paradigms on two scenes respectively from the synthetic dataset (\textit{``ROVER''}) and the real-captured dataset (\textit{``SAKANA''}).
  }
  \label{fig:teaser}
\end{figure*}

This paper investigates a novel perspective on this question by leveraging a multi-view, 3D-aware neural implicit model for efficient RGB–NIR dark imaging without relying on clean RGB supervision. Building upon the volume rendering framework and the analysis of several intuitive baseline architectures, a 3D-aware neural implicit fusion architecture is carefully redesigned to fully exploit the potential of RGB-NIR modalities. In addition, an NIR-modulated positional encoding mechanism is introduced to address the inherent limitations of conventional positional encoding under severe noise, effectively suppressing noise-induced overfitting. Finally, a Color Code MLP is presented to resolve the fundamentally ill-posed mapping between NIR and RGB through a learned color code distribution.
Comprehensive evaluations are conducted on both synthetic and real-world multi-view datasets containing NIR images and noisy RGB observations. The proposed framework demonstrates strong performance under extreme low-light conditions compared with alternative approaches. Moreover, it generalizes across varying noise levels compared to existing methods, without using the clean RGB for supervision. Codes and data will be released to support future research. The key contributions of this work are: 1) A new 3D-aware fusion model is proposed for RGB-NIR dark imaging. Different from existing RGB-NIR models, it effectively fuses noisy RGB observation with NIR cues in 3D space without requiring clean RGB supervision. 2) Based on the characteristic of RGB-NIR modalities, two insightful components are introduced to unlock the complementary potentials, further improving performance. 3) Both synthetic and real-world datasets are contributed, demonstrating the superiority of the proposed model across various scenarios.

\section{Related Work}

\noindent \textbf{Low-light photography.}
Early low-light imaging methods rely on handcrafted techniques~\cite{dong2012nonlocally,xu2018trilateral,dong2012nonlocal}, while recent methods~\cite{zamir2021multi,anwar2019real,chen2022simple,wang2022uformer} achieve superior performance with deep learning. For example, DnCNN~\cite{zamir2021multi} employs CNN for denoising, while transformer-based methods like Restormer~\cite{zamir2022restormer} further improve the quality.

\noindent \textbf{Denoising with multi-view models.}
Since the introduction of Neural Radiance Fields (NeRFs)~\cite{mildenhall2021nerf,niu2024rs,zhan2024kfd}, several studies~\cite{mildenhall2022nerf,pearl2022nan,wang2023lighting} have investigated their potential for image denoising. RawNeRF~\cite{mildenhall2022nerf} performs novel-view synthesis directly on RAW data. LLNeRF~\cite{wang2023lighting} explores denoising in the sRGB domain. Despite these advancements, leveraging NIR for robust multi-view low-light imaging remains unexplored.

\noindent \textbf{Guided image denoising.}
Beyond approaches that rely solely on RGB information for denoising, several studies~\cite{eisemann2004flash,krishnan2009dark,petschnigg2004digital,he2012guided,yan2013cross,deng2020deep,li2019joint,oh2023robust,xia2021deep,xiong2021seeing,jin2022darkvisionnet,sheng2023structure,wang2024rffnet,zhang2025sgdformer,xu2024nir} have explored the use of additional modalities, such as burst flash~\cite{eisemann2004flash,krishnan2009dark,petschnigg2004digital,he2012guided,oh2023robust,xiong2021seeing} and Near-Infrared (NIR) imaging~\cite{yan2013cross,jin2022darkvisionnet,sheng2023structure,wang2024rffnet,xu2024nir,wan2022purifying,wang2025complementary,niu2023physics}. 
SANet~\cite{sheng2023structure} estimates a clean structure map for RGB-NIR fusion. NAID~\cite{xu2024nir} further proposes a Selective Fusion Module.

\section{Approach}
\label{sec:method}

\subsection{Preliminary: MLPs and Volume Rendering}

\begin{table*}[t]
\begin{minipage}[p]{0.385\textwidth}
\centering
\setlength{\abovecaptionskip}{2mm}
\includegraphics[width=.95\textwidth]{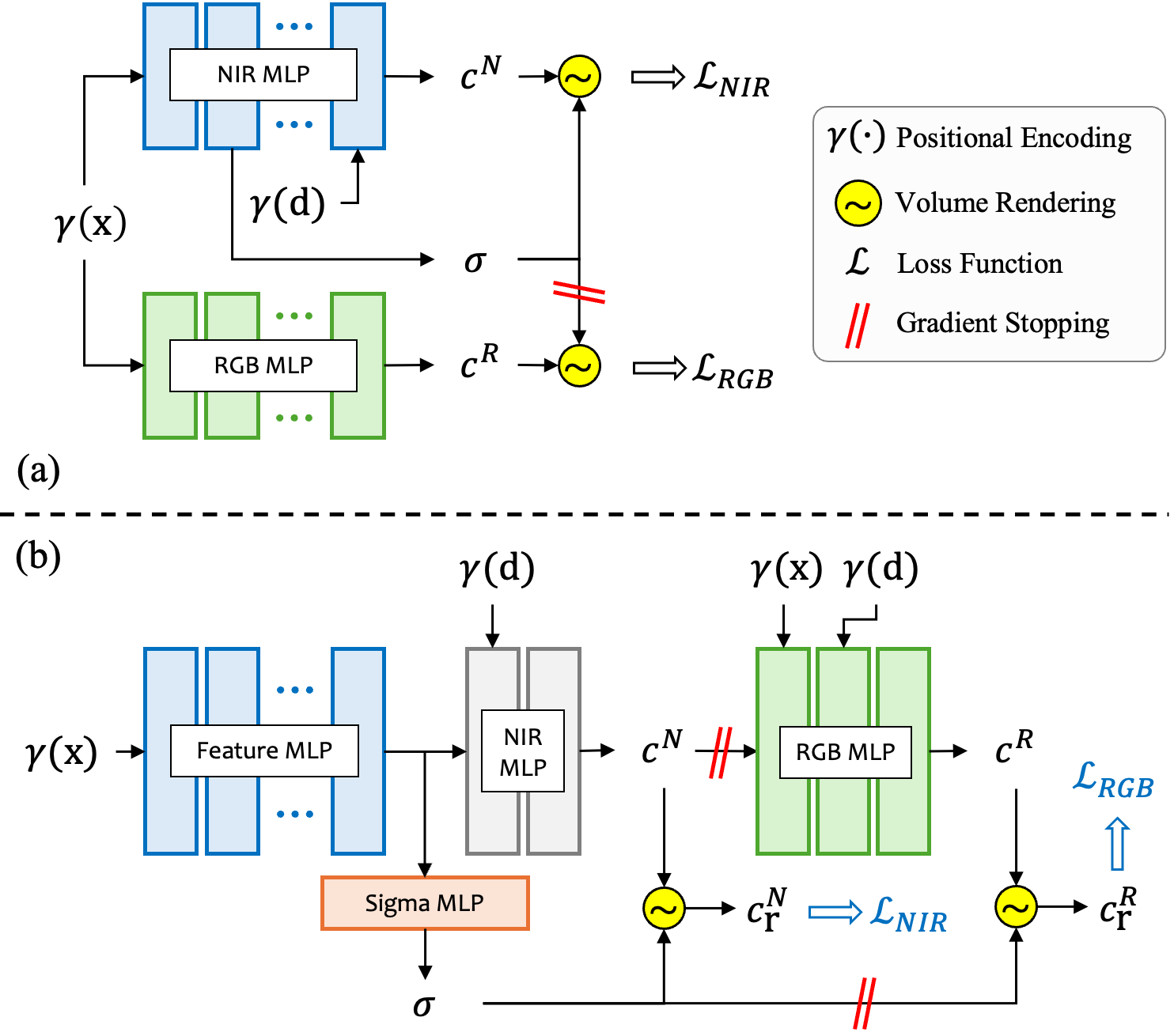}
\captionof{figure}{Architecture illustrations of (a) parallel structure and (b) NIR-conditioned structure.}
\label{fig:first_two_arch}
\end{minipage}
\hspace{0.001\textwidth}
\begin{minipage}[p]{0.605\textwidth}
\centering
\setlength{\abovecaptionskip}{0.5mm}
\includegraphics[width=.95\textwidth]{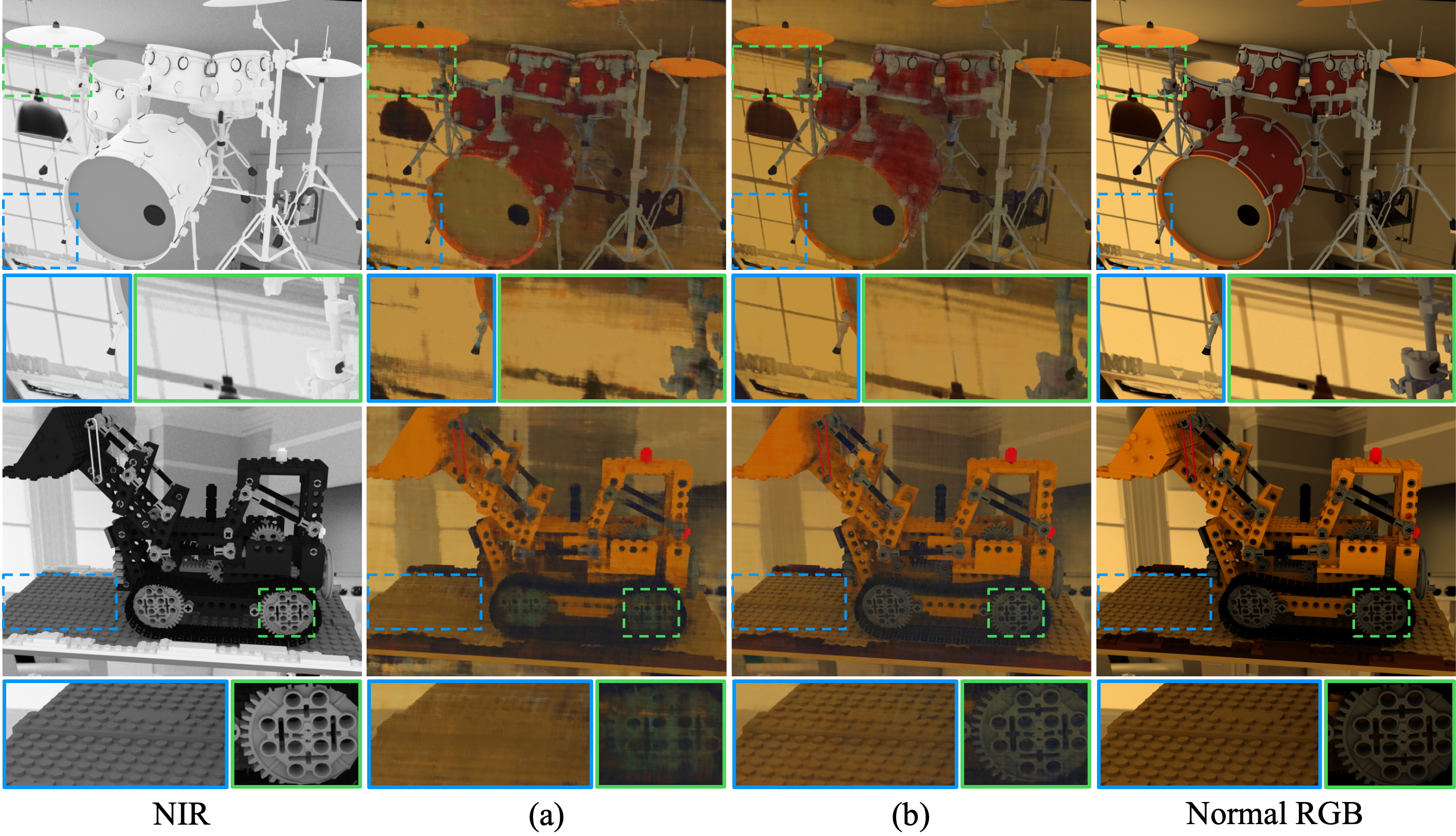}
\captionof{figure}{Visual comparison results. From left to right: NIR observation, (a) result of parallel structure, (b) result of NIR-conditioned structure, and normal RGB image.}
\label{fig:first_two_arch_result}
\end{minipage}
\end{table*}

Previous research~\cite{mildenhall2022nerf,pearl2022nan,wang2023lighting} indicates that integrating Multi-Layer Perceptrons (MLPs) with volume rendering can aid in image denoising. This ability arises from two factors: 1) MLPs inherently favor smooth, low-frequency predictions, and 2) the volume-rendering process enforces 3D consistency by aggregating information across multiple viewpoints. To validate this behavior, a vanilla NeRF architecture from~\cite{mildenhall2022nerf}
was trained using noisy RGB inputs. The results (Fig.~\ref{fig:vanilla_nerf_result}) show that although the method reduces noise to some extent, it introduces significant artifacts.
While the volume-rendering framework remains robust and crucial with the enforcement of multi-view consistency, the characteristics of MLPs need to be carefully handled to fully exploit the potentials of RGB–NIR signals, particularly when supervised by noisy RGB data. The following sections retain the standard volume-rendering formulation without modification, while introducing a completely redesigned MLP architecture specifically tailored to the complementary properties of RGB–NIR modalities.

\subsection{Fusing NIR and RGB in 3D Space}

\begin{figure}[t]
    \centering
    \setlength{\abovecaptionskip}{2mm}
    \includegraphics[width=\linewidth]{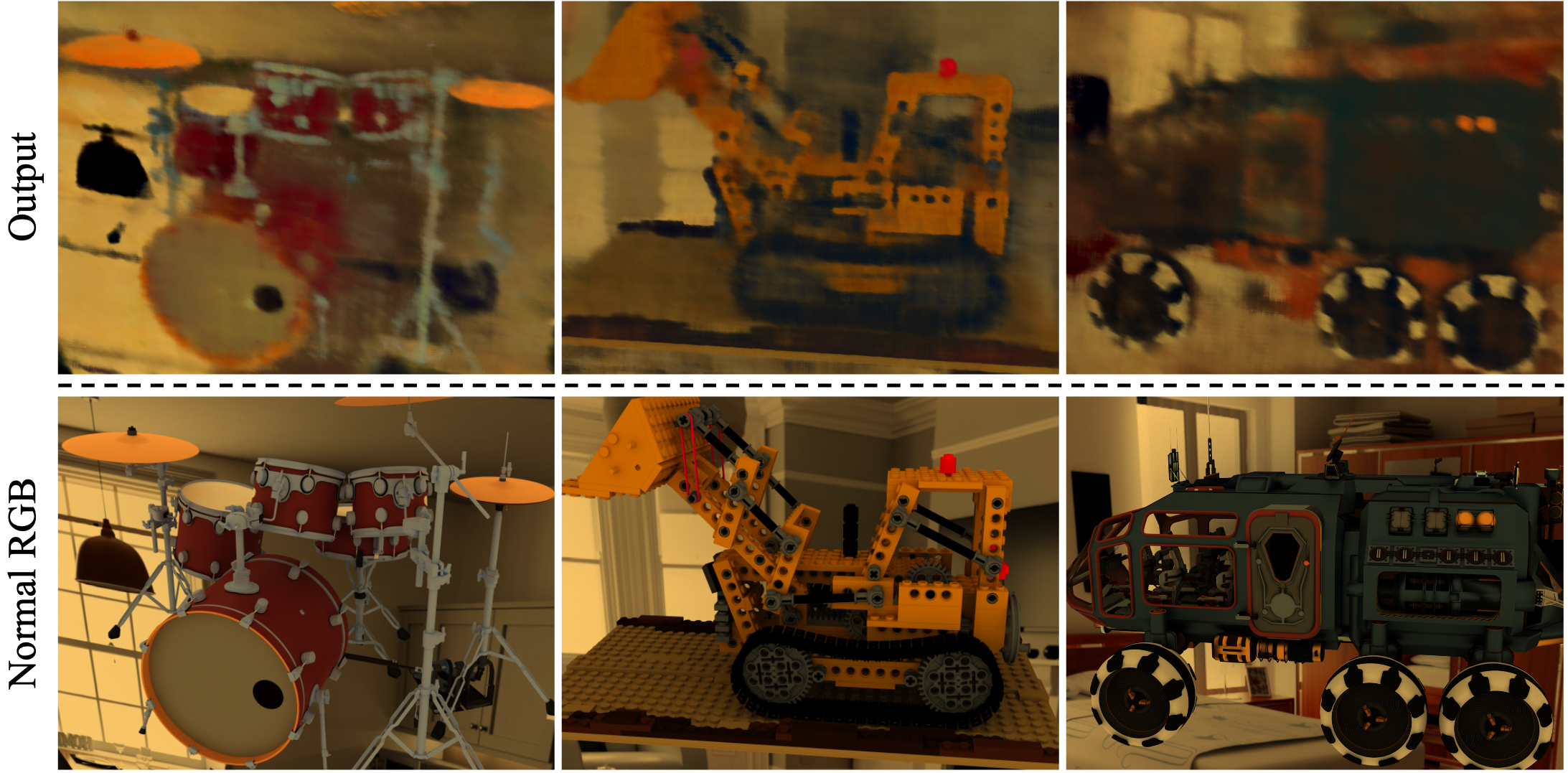}
    \caption{Results of using vanilla NeRF MLP for our model. 
    }
    \label{fig:vanilla_nerf_result}
\end{figure}

\noindent \textbf{Intuitive parallel structure.} A straightforward parallel architecture jointly optimizes two separate MLPs for RGB and NIR observations. Since the primary purpose of incorporating NIR images is to provide structural guidance, the NIR MLP is used to estimate the volume density $\sigma$, while a gradient-stopping strategy is applied to prevent degraded RGB observations from adversely affecting its learning process. The architectural design and corresponding qualitative results are presented in Fig.~\ref{fig:first_two_arch}~(a) and Fig.~\ref{fig:first_two_arch_result}~(a), respectively.
The additional structural information from NIR substantially improves the reconstruction of overall scene geometry compared with the vanilla NeRF MLP (Fig.~\ref{fig:vanilla_nerf_result}). However, the model still fails to recover detailed 2D texture information already present in the NIR images, such as the background texture and the floor in the ``Lego'' scene. This limitation arises because the RGB MLP does not directly exploit the predicted NIR values, which encode rich texture information crucial for high-fidelity reconstruction.

\begin{table*}[t]
\begin{minipage}[p]{0.345\textwidth}
\centering
\setlength{\abovecaptionskip}{2mm}
\includegraphics[width=\textwidth]{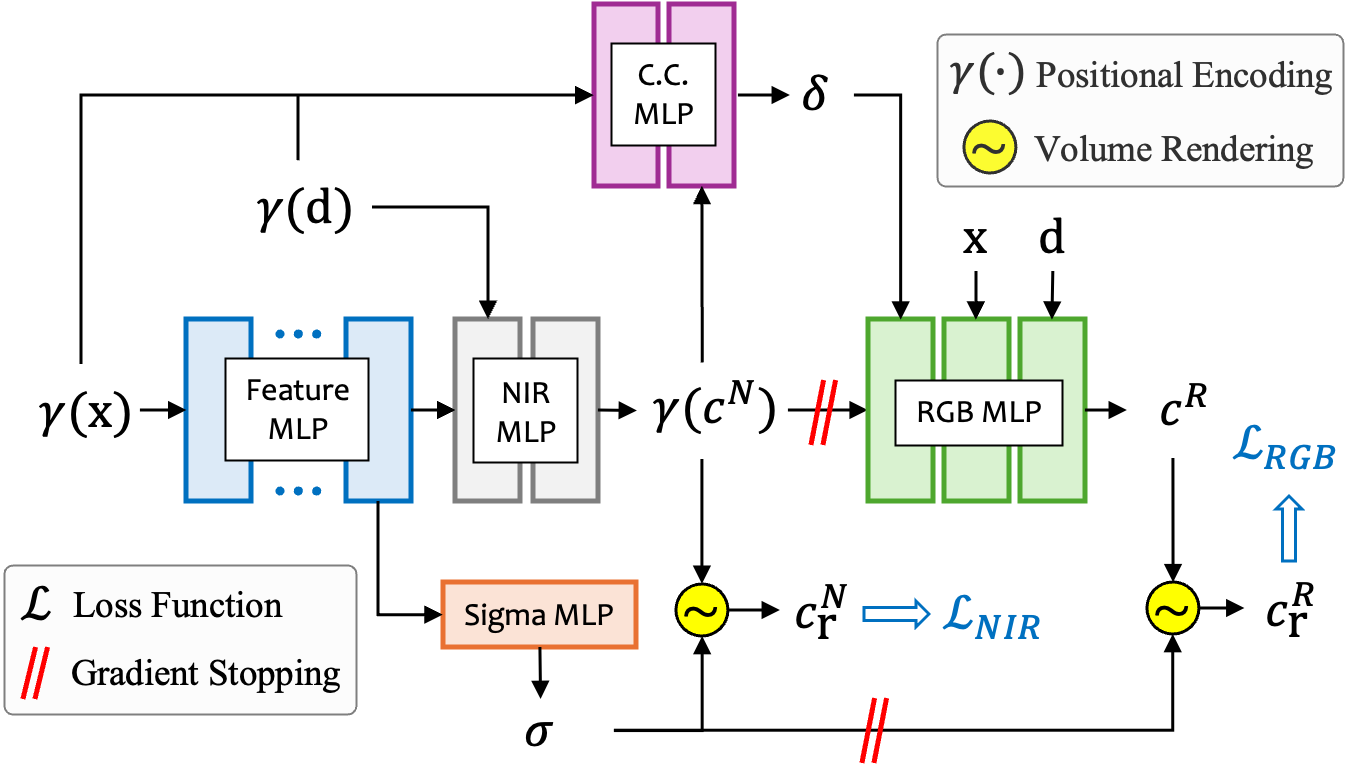}
\captionof{figure}{Illustration of the final architecture with the Color Code MLP.}
\label{fig:cce_arch}
\end{minipage}
\hspace{0.001\textwidth}
\begin{minipage}[p]{0.645\textwidth}
\centering
\setlength{\abovecaptionskip}{0.5mm}
\includegraphics[width=\textwidth]{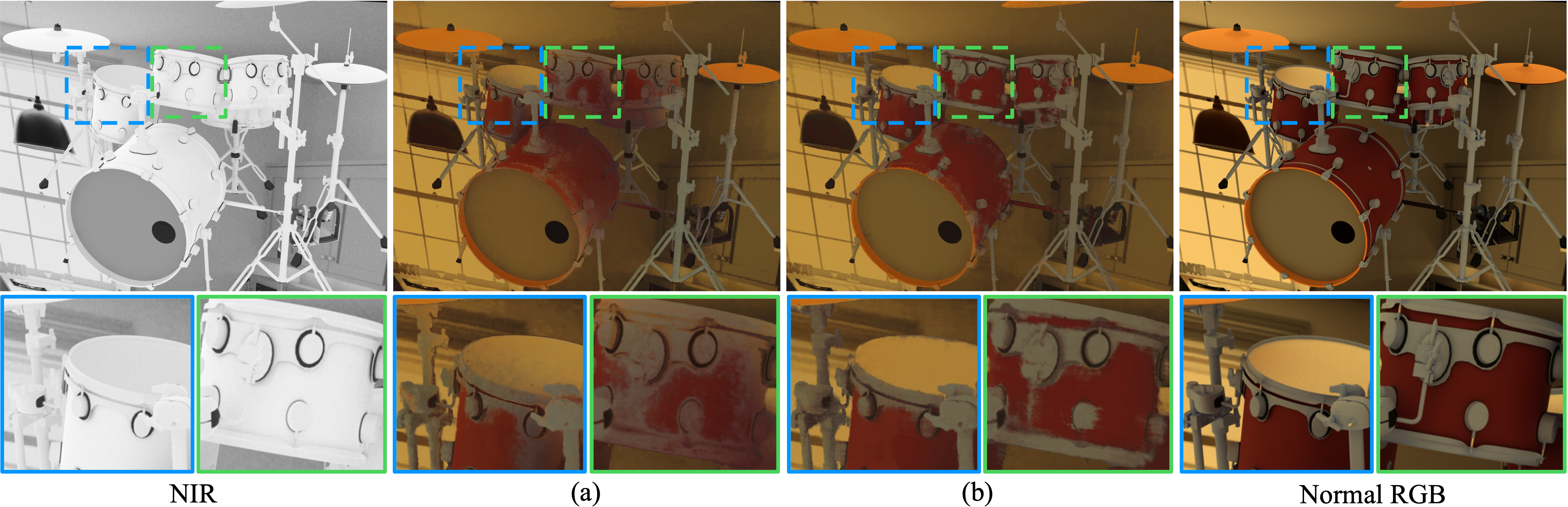}
\captionof{figure}{Visual comparison results. From left to right: NIR observation, (a) result without Color Code MLP, (b) result with Color Code MLP, and normal RGB image.}
\label{fig:nir_rgb_gap}
\end{minipage}
\end{table*}

\begin{figure}[t]
\centering
\setlength{\abovecaptionskip}{2mm}
\includegraphics[width=.9\linewidth]{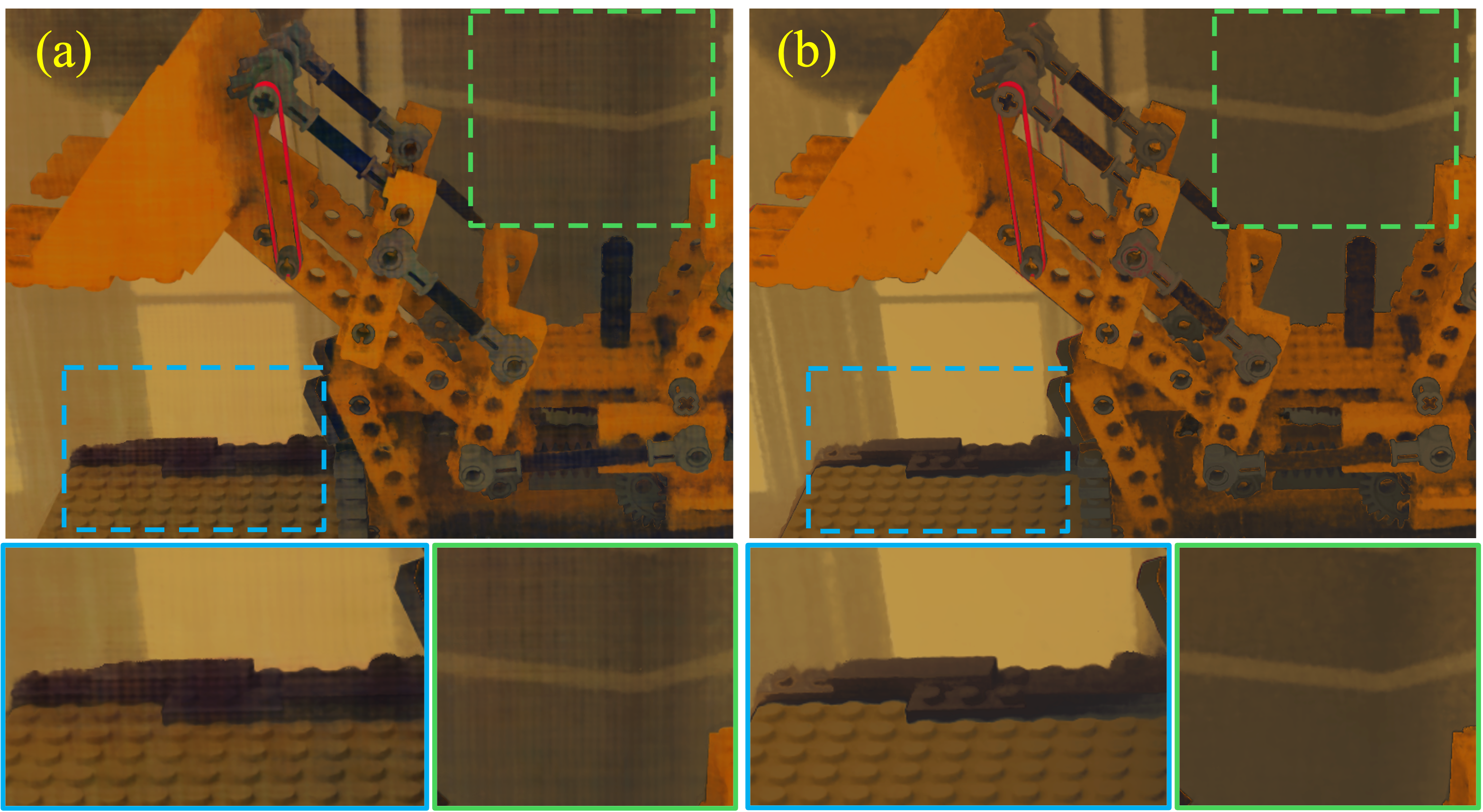}
\caption{
(a) Applying P.E. to \(\mathbf{x}\) and \(\mathbf{d}\) in the RGB MLP introduces checkerboard artifacts.
(b) Applying P.E. to \(\mathbf{c}^{N}\) instead effectively removes these artifacts and enhances the reconstruction of structural details.
}
\label{fig:checkerboard}
\end{figure}

\noindent \textbf{NIR-conditioned structure.} Building on the previous observations, the architecture is improved by feeding the NIR output into the RGB MLP while ensuring that gradients from the RGB MLP do not propagate back to the NIR MLP. This modification enables the RGB MLP to leverage the structural guidance provided by NIR while maintaining the integrity of the NIR representation. The revised architecture and its corresponding qualitative results are shown in Fig.~\ref{fig:first_two_arch}~(b) and Fig.~\ref{fig:first_two_arch_result}~(b), respectively.
This enhanced design substantially improves both 3D structural reconstruction and the recovery of fine 2D texture details. By implicitly fusing NIR predictions with RGB predictions in 3D space, the model effectively exploits additional structural information for improved results.

\subsection{Revisiting Positional Encoding with NIR}
\label{sec:nir_pe}

Positional Encoding (P.E.) transforms 3D coordinates $\mathbf{x}$ and 2D view directions $\mathbf{d}$ into a higher-dimensional representation using sinusoidal bases, enabling the network to model high-frequency details. However, under noisy conditions where most high-frequency variations are noise, noticeable ``checkerboard effects'' are observed in the rendering results (Fig.~\ref{fig:checkerboard}~(a)). This issue arises because P.E. \emph{universally} assumes that high-frequency details can occur at any position $\mathbf{x}$, an assumption valid for clean RGB supervision but leading to \emph{overfitting on noise} when optimized on noisy RGB.

\begin{figure}[t]
\centering
\setlength{\abovecaptionskip}{1mm}
\includegraphics[width=\linewidth]{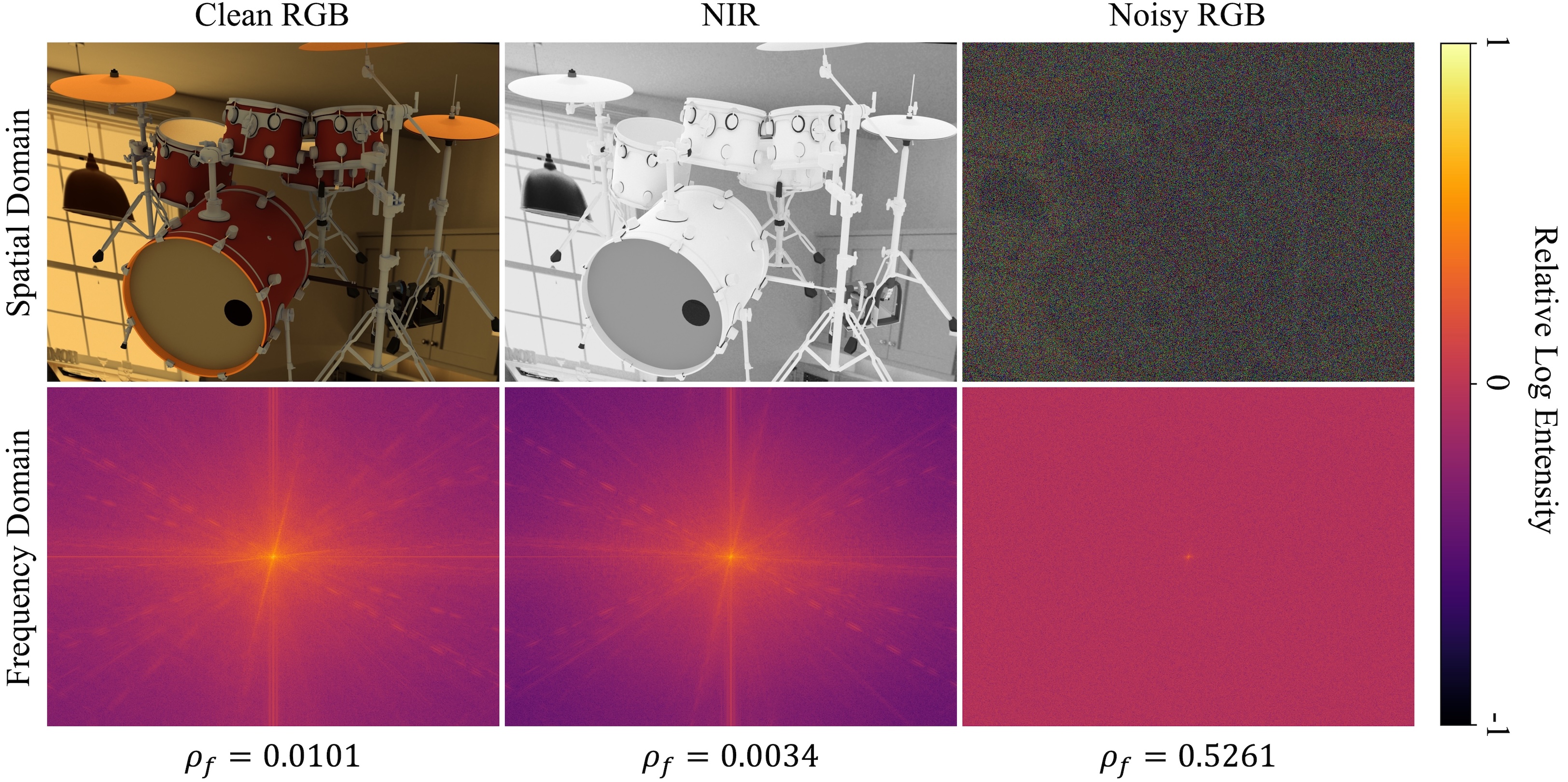}
\caption{Clean RGB, NIR, and noisy RGB in frequency domain. $\rho_{f}$ represents the energy ratio of high/low frequency.
}
\label{fig:NIR_PE_frequency}
\end{figure}

Fig.~\ref{fig:NIR_PE_frequency} visualizes the frequency-domain distributions of clean RGB, NIR, and noisy RGB. The clean RGB and NIR show similar frequency distributions indicating smooth, natural structures, whereas the noisy RGB exhibits strong dispersion across the entire frequency range dominated by random high-frequency noise. This observation is quantified by the high-to-low frequency energy ratio $\rho_{f}$. The noisy RGB exhibits a much larger ratio ($0.5261$), while clean RGB and NIR are significantly lower ($0.0101$ and $0.0034$, respectively), indicating that noisy RGB is dominated by high-frequency noise. This suggests that under the supervision of extremely noisy RGB, traditional positional encoding is prone to overfit meaningless high-frequency noise, resulting in ``checkerboard effects.''

Here the insight is to fully utilize the similarity between NIR and RGB in the frequency domain. Fig.~\ref{fig:NIR_PE_frequency} suggests that clean RGB possesses a similar frequency distribution to NIR, with slightly richer high-frequency content due to color/texture variations. Based on this analysis, instead of applying P.E. to $(\mathbf{x},\mathbf{d})$, the encoding is applied to the NIR estimation $\mathbf{c}^{N}$, while $(\mathbf{x},\mathbf{d})$ are directly fed into the RGB MLP. Concretely, let $\gamma(\cdot)$ denote the standard sinusoidal positional encoding; the RGB MLP takes $\mathbf{x}$, $\mathbf{d}$, and $\gamma(\mathbf{c}^{N})$ as input and predicts the RGB color $\mathbf{c}^{R}$:
\begin{equation}
\label{eq:nir-pe}
\mathbf{c}^{R} = \operatorname{RGB MLP}\bigl(\,\mathbf{x},\,\mathbf{d},\,\gamma(\mathbf{c}^{N})\,\bigr).
\end{equation}
This strategy acts as a \emph{frequency-aligned modulation} that amplifies structurally consistent frequencies present in NIR while suppressing noise-driven high-frequency components from the noisy RGB supervision. Fig.~\ref{fig:checkerboard}~(b) shows that this crucial modification effectively eliminates checkerboard artifacts and improves structural reconstruction.

\subsection{Resolving the RGB-NIR Ambiguities}

Despite the strong structural guidance provided by NIR, several ``blending'' effects appear in the rendering results (Fig.~\ref{fig:nir_rgb_gap}~(a)), hindering accurate recovery of RGB colors and texture details. This issue arises from the inherent NIR-to-RGB color ambiguity, where a single NIR value can correspond to multiple RGB values.
As illustrated in the zoomed-in region of Fig.~\ref{fig:nir_rgb_gap}, points with identical NIR values may exhibit distinct RGB colors, making the learning process ill-posed for MLPs. Consequently, the network fails to distinguish these variations and instead produces blended RGB estimates for points sharing the same NIR value (Fig.~\ref{fig:nir_rgb_gap}~(a)). Although the RGB MLP takes 3D coordinates \(\mathbf{x}\) and 2D view directions \(\mathbf{d}\) as additional inputs, the smoothness of MLPs prevents the model from producing high-frequency RGB estimates for spatially adjacent points with similar NIR values, resulting in soft transition artifacts.

\begin{figure}[t]
\centering
\setlength{\abovecaptionskip}{2mm}
\includegraphics[width=\linewidth]{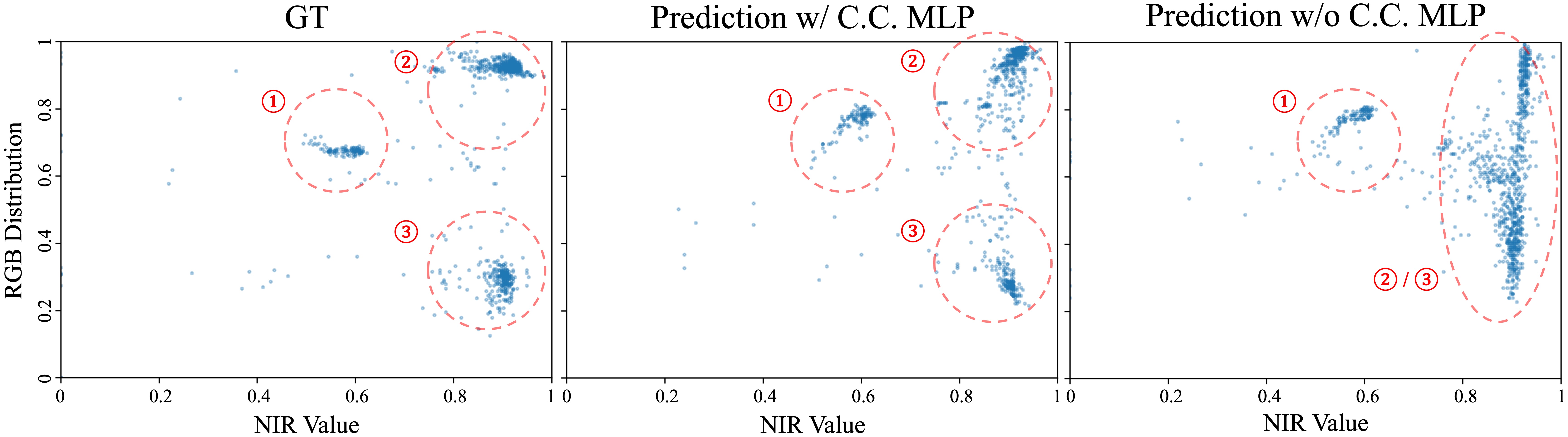}
\caption{Distribution of RGB regarding to the NIR value.
}
\label{fig:ccmlp_distribution}
\end{figure}

\begin{figure*}[t]
\centering
\setlength{\abovecaptionskip}{2mm}
\includegraphics[width=\linewidth]{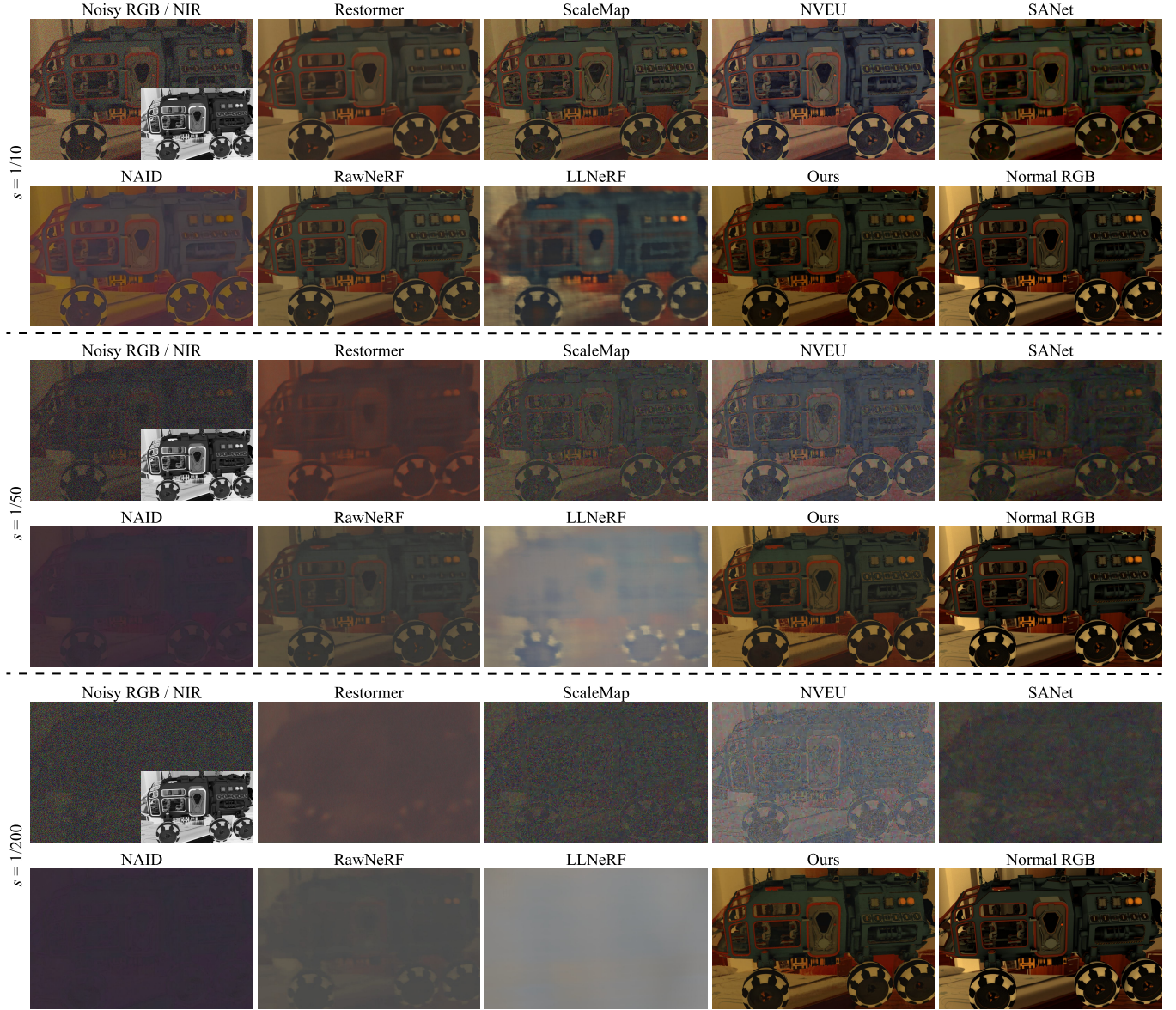}
\caption{Qualitative comparison results on synthetic dataset. 
}
\label{fig:comparison}
\end{figure*}

\begin{table*}[t]
\centering
\begin{minipage}{\textwidth}
\centering
\setlength{\abovecaptionskip}{2mm}
\includegraphics[width=\linewidth]{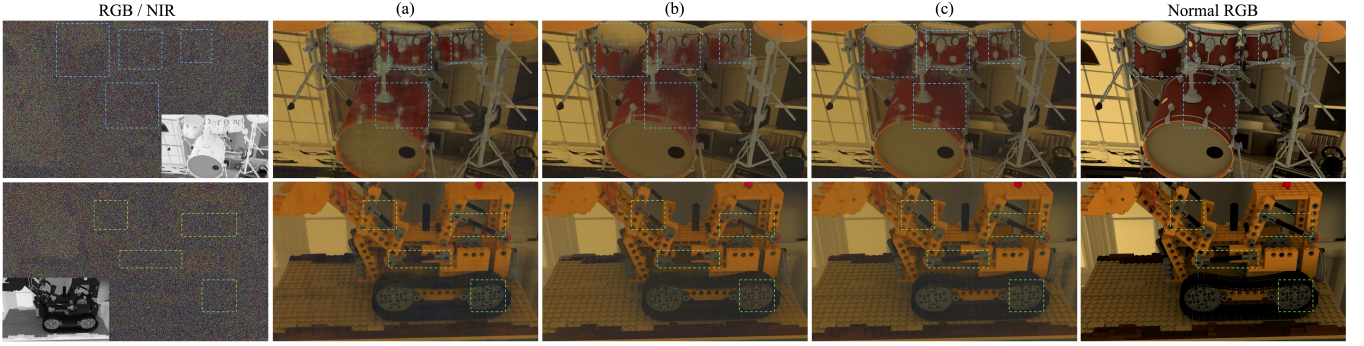}
\captionof{figure}{Qualitative ablation results on synthetic dataset. (a) Result w/o NIR-P.E. (b) Result w/o C.C. MLP. (c) Result with full components. Zoom in for the best view. 
}
\label{fig:ablation}
\vspace{2mm}
\end{minipage}
\begin{minipage}{\textwidth}
\centering
\setlength{\abovecaptionskip}{2mm}
\caption{Quantitative comparison results on synthetic dataset.
Best and second best results are annotated with bold and underline.
}
\label{tab:comparison}
\resizebox{\textwidth}{!}{%
\begin{tabular}{lccccccccccccccc}
\toprule
\multirow{2}{*}{Methods} & \multicolumn{3}{c}{$s=1/10$} & \multicolumn{3}{c}{$s=1/25$} & \multicolumn{3}{c}{$s=1/50$} & \multicolumn{3}{c}{$s=1/100$} & \multicolumn{3}{c}{$s=1/200$} \\
 & SSIM $\uparrow$ & PSNR $\uparrow$ & LPIPS $\downarrow$ & SSIM $\uparrow$ & PSNR $\uparrow$ & LPIPS $\downarrow$ & SSIM $\uparrow$ & PSNR $\uparrow$ & LPIPS $\downarrow$ & SSIM $\uparrow$ & PSNR $\uparrow$ & LPIPS $\downarrow$ & SSIM $\uparrow$ & PSNR $\uparrow$ & LPIPS $\downarrow$ \\ \midrule
Restormer & 0.6577 & 17.60 & 0.3561 & 0.5886 & \underline{16.29} & 0.5397 & 0.5489 & \underline{15.43} & 0.6779 & 0.4993 & \underline{14.91} & 0.8265 & 0.4941 & \underline{14.43} & 0.8245 \\
ScaleMap & 0.6902 & 17.49 & 0.2793 & 0.5874 & 15.98 & 0.4611 & 0.4973 & 15.10 & 0.6552 & 0.3749 & 14.50 & 0.8393 & 0.2873 & 13.91 & 0.9317 \\
NVEU & 0.4586 & 15.44 & 0.5025 & 0.4373 & 15.07 & 0.5938 & 0.4055 & 14.06 & 0.6709 & 0.3552 & 12.38 & \underline{0.7253} & 0.3108 & 11.26 & \underline{0.7596} \\
SANet & \underline{0.7556} & 17.52 & 0.2966 & \underline{0.6431} & 15.53 & 0.4376 & 0.4934 & 14.62 & 0.6785 & 0.2366 & 14.21 & 1.1068 & 0.1592 & 13.64 & 1.2321 \\
NAID & 0.5351 & 16.13 & 0.3939 & 0.5223 & 14.85 & 0.6353 & 0.5061 & 14.17 & 0.7686 & 0.4973 & 13.80 & 0.8144 & 0.4929 & 13.61 & 0.8257 \\
RawNeRF & 0.7047 & 17.49 & \underline{0.2384} & 0.6244 & 15.99 & \underline{0.3678} & \underline{0.5730} & 15.15 & \underline{0.5391} & \underline{0.5371} & 14.67 & 0.7377 & \underline{0.5157} & 14.21 & 0.8651 \\
LLNeRF & 0.5844 & \underline{17.82} & 0.6202 & 0.5023 & 13.96 & 0.7477 & 0.4569 & 11.87 & 0.8321 & 0.4240 & 10.12 & 0.8910 & 0.4045 & 9.01 & 0.9193 \\
Ours & \textbf{0.7572} & \textbf{19.53} & \textbf{0.2361} & \textbf{0.7441} & \textbf{19.52} & \textbf{0.2354} & \textbf{0.7485} & \textbf{19.70} & \textbf{0.2365} & \textbf{0.7299} & \textbf{19.87} & \textbf{0.2485} & \textbf{0.7028} & \textbf{19.68} & \textbf{0.2670} \\ \bottomrule
\end{tabular}%
}
\end{minipage}
\end{table*}

To address this issue, a Color Code MLP (C.C. MLP) is introduced to learn a color distribution for each NIR value \(\mathbf{c}^{N}\). Specifically, the C.C. MLP predicts a non-uniform log probability distribution \( \pi \), which is used to obtain the one-hot color code \( \delta \):
\begin{equation}
\delta = \operatorname{one\_hot} \left( \arg\max_{i} \left[ g_i + \log \pi_i \right] \right),
\end{equation}
where Gumbel-Softmax~\cite{jang2016categorical} is introduced to maintain differentiability. \( g_1, \ldots, g_k \) are i.i.d. samples from the $\operatorname{Gumbel}(0, 1)$ distribution~\cite{gumbel1954statistical,maddison2014sampling,jang2016categorical}, and \( \pi_i \) denotes the predicted log probabilities for each of the \( k \) categories. The final output of the C.C. MLP is a \( K \)-dimensional one-hot vector \( \delta \), which is then used as input to the RGB MLP for color prediction. $K$ is set to $16$ by default. This design enables the model to resolve NIR-to-RGB ambiguity by learning a probabilistic mapping from NIR values to plausible RGB variations. As illustrated in Fig.~\ref{fig:nir_rgb_gap}, the incorporation of the C.C. MLP significantly improves rendering quality by addressing the fundamental NIR-to-RGB ambiguity.

To further demonstrate the effectiveness of the proposed C.C. MLP, the RGB feature distributions of the GT, the prediction with C.C. MLP, and the prediction without C.C. MLP are visualized in Fig.~\ref{fig:ccmlp_distribution}. The RGB values are projected onto a 1D space using PCA, and $1000$ points are randomly sampled for comparison. The GT distribution exhibits three separate clusters corresponding to distinct RGB colors (circled in \ding{172}, \ding{173}, and \ding{174}). While the prediction without C.C. MLP correctly estimates the RGB distribution within \ding{172}, it fails to distinguish clusters \ding{173} and \ding{174}, as they share similar NIR values. In contrast, the model equipped with C.C. MLP successfully separates these clusters, indicating its ability to resolve the NIR-to-RGB ambiguity and preserve diverse color representations. Fig.~\ref{fig:cce_arch} shows the final architecture.

\subsection{Optimization}
\label{sec:optimization}

\noindent \textbf{Loss function.}
All MLPs are jointly optimized by minimizing the L2 photometric loss between the rendered NIR estimation \(\mathbf{c}_{\mathbf{r}}^{N}\) and RGB estimation \(\mathbf{c}_{\mathbf{r}}^{R}\) and their corresponding NIR observation \(\hat{\mathbf{c}}_{\mathbf{r}}^{N}\) and noisy RGB observation \(\hat{\mathbf{c}}_{\mathbf{r}}^{R}\):
\begin{equation}
\mathcal{L}_{photo} = ||\mathbf{c}_{\mathbf{r}}^{N} - \hat{\mathbf{c}}_{\mathbf{r}}^{N}||_2 + ||\mathbf{c}_{\mathbf{r}}^{R} - \left(\hat{\mathbf{c}}_{\mathbf{r}}^{R}\right)^{\frac{1}{\phi}}||_2,
\end{equation}
where $\phi$ is a hyperparameter to amplify the exposure of $\hat{\mathbf{c}}_{\mathbf{r}}^{R}$.

\noindent \textbf{Training details.}
The model is optimized for \(200,000\) iterations using the Adam optimizer~\cite{kingma2014adam} with a learning rate of \(5 \times 10^{-4}\), \(\beta_1 = 0.9\), and \(\beta_2 = 0.999\). The training process requires approximately 3 hours on one NVIDIA RTX 4090 GPU, with a peak memory consumption of 9 GB.

\section{Evaluations}
\label{sec:experiment}

\subsection{Benchmarks}
\label{sec:dataset}

\begin{table*}[t]
    \centering
    \begin{minipage}{\textwidth}
        \centering
        \setlength{\abovecaptionskip}{2mm}
        \includegraphics[width=\linewidth]{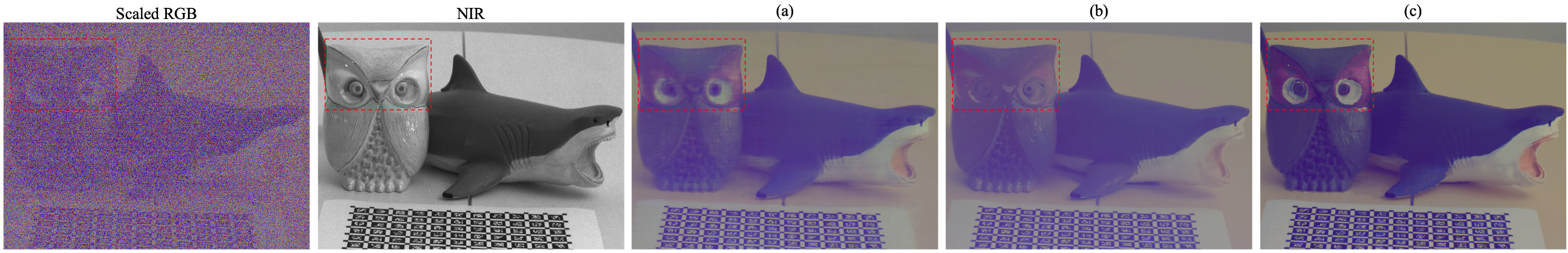}
        \captionof{figure}{Qualitative ablation results on real-world captures. 
        (a) Result w/o NIR-P.E. (b) Result w/o C.C. MLP. (c) Result with full components. Zoom in for the best view. 
        }
        \label{fig:real_ablation}
    \vspace{3mm}
    \end{minipage}
    \begin{minipage}{\textwidth}
        \centering
        \setlength{\abovecaptionskip}{2mm}
        \includegraphics[width=\linewidth]{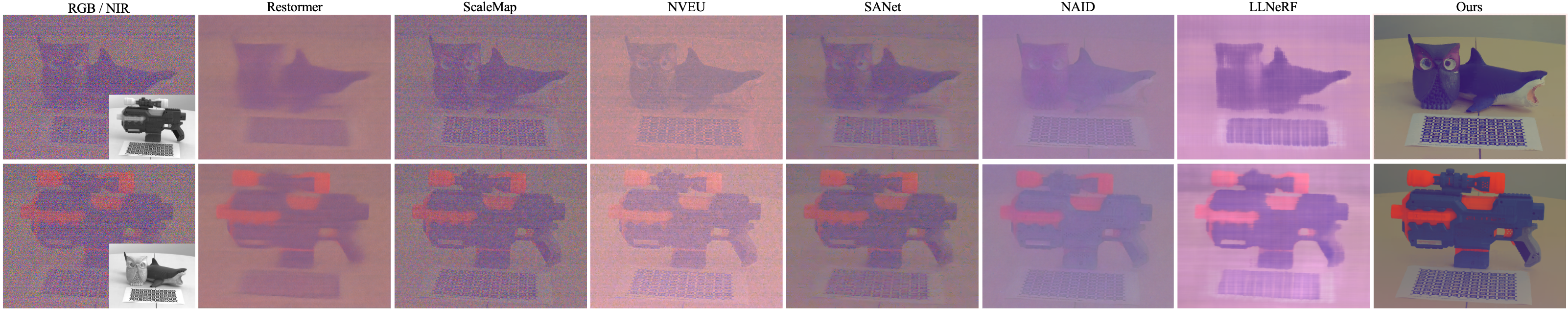}
        \captionof{figure}{Visual comparisons on real-world captures. Zoom in for the best view. 
        }
        \label{fig:real_comparison}
    \end{minipage}
\end{table*}

\noindent \textbf{Synthetic data.}
Evaluating the proposed method requires a multi-view dataset containing low-light RGB and corresponding NIR images. Since no such dataset is publicly available, a synthetic dataset is generated using Mitsuba3~\cite{Mitsuba3}. Each scene consists of $40$ randomly sampled camera views, rendering both a normal RGB image and a \SI{850}{nm} NIR image at a resolution of $960 \times 1280$. To simulate noisy RGB images, low-light conditions are created by scaling the RGB pixel values by $s$. Physics-based shot noise and read noise are then synthesized following~\cite{brooks2019unprocessing}. The dataset contains $5$ scenes, each with $5$ scale factors $s \in \{1/10, 1/25, 1/50, 1/100, 1/200\}$ representing different noise levels.
For each scene, $5$ views are reserved for testing, and the remaining $35$ views are used for training.

\begin{table}[t]
\centering
\setlength{\abovecaptionskip}{2mm}
\caption{Quantitative ablation results on synthetic dataset. 
}
\label{tab:ablation}
\resizebox{\linewidth}{!}{
\begin{tabular}{llccccc}
\toprule
Methods & Metrics & \small{$s\!=\!1/10$} & \small{$s\!=\!1/25$} & \small{$s\!=\!1/50$} & \small{$s\!=\!1/100$} & \small{$s\!=\!1/200$} \\ \midrule
\multirow{3}{*}{w/o NIR-P.E.} & SSIM $\uparrow$ & \underline{0.7563} & \underline{0.7365} & \underline{0.7476} & \underline{0.7292} & \underline{0.6936} \\
 & PSNR $\uparrow$ & \underline{19.48} & \underline{19.49} & \underline{19.62} & \underline{19.75} & \textbf{19.69} \\
 & LPIPS $\downarrow$ & \underline{0.2378} & \underline{0.2360} & \underline{0.2438} & 0.2657 & 0.3072 \\ \midrule
\multirow{3}{*}{w/o C.C. MLP} & SSIM $\uparrow$ & 0.7428 & 0.7294 & 0.7324 & 0.7153 & 0.6923 \\
 & PSNR $\uparrow$ & 19.34 & 19.25 & 19.42 & 19.61 & 19.42 \\
 & LPIPS $\downarrow$ & 0.2563 & 0.2573 & 0.2566 & \underline{0.2614} & \underline{0.2775} \\ \midrule
\multirow{3}{*}{Ours} & SSIM $\uparrow$ & \textbf{0.7572} & \textbf{0.7441} & \textbf{0.7485} & \textbf{0.7299} & \textbf{0.7028} \\
 & PSNR $\uparrow$ & \textbf{19.53} & \textbf{19.52} & \textbf{19.70} & \textbf{19.87} & \underline{19.68} \\
 & LPIPS $\downarrow$ & \textbf{0.2361} & \textbf{0.2354} & \textbf{0.2365} & \textbf{0.2485} & \textbf{0.2670} \\
\bottomrule
\end{tabular}%
}
\end{table}

\noindent \textbf{Real-world data.}
To further evaluate the applicability of the proposed method in real-world scenarios, $4$ real-world scenes are captured using a JAI FS-3200T10GE-NNC camera under low-light conditions. Each scene contains $49$ views with a resolution of $768 \times 1024$.
An \SI{850}{nm} NIR LED is used for illumination. For each scene, $5$ views are reserved for testing, and the remaining $44$ views are used for training. Comparisons are first performed in sRGB space against various SOTA baselines. Raw data is converted into 8-bit sRGB space using RawPy. Evaluations are then conducted in 16-bit RAW space, demonstrating the generality of the proposed method. For results on sRGB space, manual color correction are applied on the output for visualization. All results are reported based on the test views. COLMAP~\cite{schoenberger2016sfm} is used to calibrate the camera poses.

\subsection{Synthetic Data Evaluation}

\noindent \textbf{Comparison.} Although this study focuses on extremely noisy scenarios under low-light conditions, synthetic experiments are conducted across multiple noise levels, ranging from moderate to severe noise interference in extreme darkness. 
The proposed model is compared with SOTAs from different domains, covering both supervised and unsupervised models: Restormer~\cite{zamir2022restormer} trains a transformer for RGB image denoising. ScaleMap~\cite{yan2013cross} performs RGB-NIR denoising with hand-crafted operators. NVEU~\cite{niu2023nir} leverages large-scale unpaired clean RGB to train an RGB-NIR dark denoising model in an unsupervised way.
SANet~\cite{sheng2023structure} and NAID~\cite{xu2024nir} are supervised models for NIR–RGB dark denoising. RawNeRF~\cite{mildenhall2022nerf} enhances multi-view RAW images. LLNeRF~\cite{wang2023lighting} is designed for multi-view sRGB low-light enhancement.
The results are reported in Tab.~\ref{tab:comparison} and Fig.~\ref{fig:comparison}.
The proposed model consistently produces accurate results without using clean RGB as supervision, even in the most challenging scenarios.

In addition, a two-stage baseline is implemented where NIR and RGB images are first fused using a 2D fusion method such as SANet~\cite{sheng2023structure}, NAID~\cite{xu2024nir}, and NVEU~\cite{niu2023nir}, followed by training a NeRF on the fused results. The quantitative results are reported in Tab.~\ref{tab:two_stage}.
The proposed model achieves superior performance across different noise levels.
Qualitative comparisons are provided in the supplementary, showing that the two-stage pipeline struggles to restore accurate RGB, whereas the proposed method maintains high fidelity. This performance gap arises because 2D fusion models cannot exploit multi-view consistency when combining RGB and NIR modalities. The proposed approach achieves robustness across different noise levels without using clean RGB for optimization.

\noindent \textbf{Ablation Study.}
In addition to the analysis in Sec.~\ref{sec:method}, key components of the proposed model are further ablated to evaluate their effectiveness.
\textbf{w/o NIR-P.E.}: A variant that removes NIR-based positional encoding. \textbf{w/o C.C. MLP}: A variant that excludes the Color Code MLP (C.C. MLP) from the architecture.
The corresponding qualitative and quantitative results are presented in Fig.~\ref{fig:ablation} and Tab.~\ref{tab:ablation}, respectively. The results indicate that removing NIR-based positional encoding leads to pronounced ``checkerboard effects,'' resulting in visually unsatisfactory renderings. Similarly, removing the C.C. MLP causes the model to misestimate RGB values for spatially adjacent 3D points sharing identical NIR values, leading to noticeable blending artifacts.

\subsection{Real-World Data Evaluation}

\begin{table}[t]
\centering
\setlength{\abovecaptionskip}{2mm}
\setlength{\tabcolsep}{2mm}
\caption{Quantitative Results of ``2D Fusion + NeRF'' baseline under different scale factor $s$. 
}
\label{tab:two_stage}
\resizebox{\linewidth}{!}{
\begin{tabular}{llccccc}
\toprule
Methods & Metrics & \small{$s\!=\!1/10$} & \small{$s\!=\!1/25$} & \small{$s\!=\!1/50$} & \small{$s\!=\!1/100$} & \small{$s\!=\!1/200$} \\ \midrule
\multirow{3}{*}{\shortstack[l]{NVEU\\+ NeRF}} & SSIM $\uparrow$ & 0.5278 & 0.5076 & 0.4644 & 0.4183 & 0.3610 \\
 & PSNR $\uparrow$ & 16.03 & 15.71 & 14.58 & 12.89 & 11.90 \\
 & LPIPS $\downarrow$ & 0.4652 & 0.5571 & 0.6339 & 0.6946 & \underline{0.7214} \\ \midrule
\multirow{3}{*}{\shortstack[l]{SANet\\+ NeRF}} & SSIM $\uparrow$ & \underline{0.7539} & \underline{0.6553} & \underline{0.5834} & \underline{0.5468} & \underline{0.5231} \\
 & PSNR $\uparrow$ & \underline{17.54} & 15.46 & \underline{14.66} & \underline{14.63} & \underline{14.42} \\
 & LPIPS $\downarrow$ & \underline{0.2878} & \underline{0.4299} & \underline{0.5979} & \underline{0.7612} & 0.8674 \\ \midrule
\multirow{3}{*}{\shortstack[l]{NAID\\+ NeRF}} & SSIM $\uparrow$ & 0.6102 & 0.5658 & 0.4506 & 0.3871 & 0.3619 \\
 & PSNR $\uparrow$ & 16.71 & \underline{16.14} & 14.42 & 13.01 & 12.23 \\
 & LPIPS $\downarrow$ & 0.4478 & 0.6391 & 0.7793 & 0.8398 & 0.8623 \\ \midrule
\multirow{3}{*}{Ours} & SSIM $\uparrow$ & \textbf{0.7572} & \textbf{0.7441} & \textbf{0.7485} & \textbf{0.7299} & \textbf{0.7028} \\
 & PSNR $\uparrow$ & \textbf{19.53} & \textbf{19.52} & \textbf{19.70} & \textbf{19.87} & \textbf{19.68} \\
 & LPIPS $\downarrow$ & \textbf{0.2361} & \textbf{0.2354} & \textbf{0.2365} & \textbf{0.2485} & \textbf{0.2670} \\
\bottomrule
\end{tabular}%
}
\end{table}

\begin{figure*}[t]
\centering
    \centering
    \setlength{\abovecaptionskip}{2mm}
    \includegraphics[width=\linewidth]{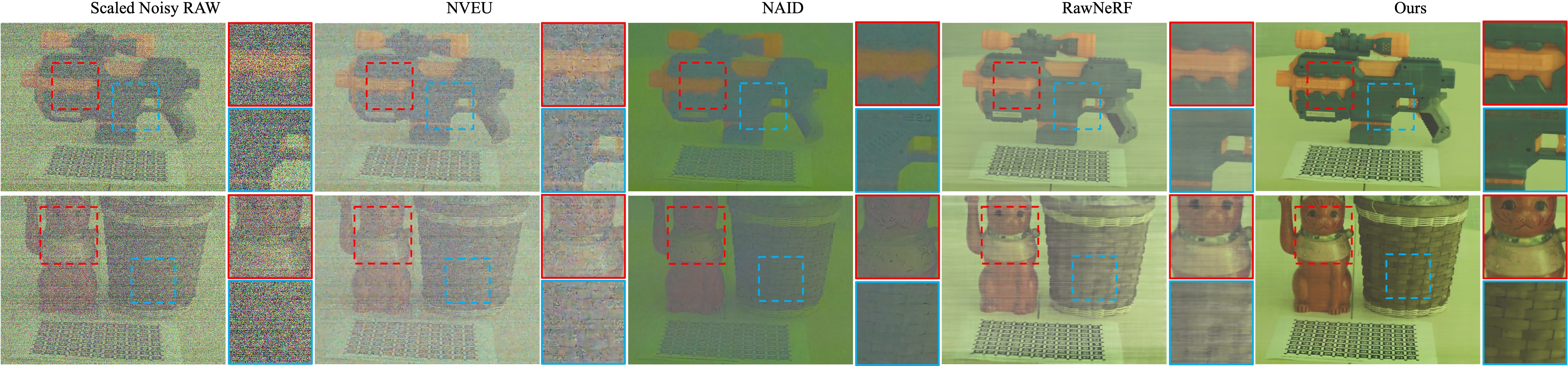}
    \caption{%
    Qualitative Comparison results in RAW space. Zoom in for the best view. 
    }
    \label{fig:rawnerf}
\end{figure*}

To evaluate the applicability of different methods, experiments are conducted on $4$ real-world scenes. For quantitative assessment, commonly used non-reference image quality metrics are employed, including Perceptual Index (PI)~\cite{gu2022ntire}, MUSIQ~\cite{ke2021musiq}, and MANIQA~\cite{yang2022maniqa}. In addition, a Human Subjective Evaluation (HSE) is conducted to directly reflect human perceptual judgment. Specifically, ten volunteers are asked to independently evaluate the results and assign an integer score from 1 (poor) to 5 (excellent).

\noindent \textbf{Ablation Study.}
The qualitative ablation results on real-world captures are presented in Fig.~\ref{fig:real_ablation}. The model exhibits ``checkerboard effects'' when NIR-P.E. is removed. Additionally, excluding the C.C. MLP leads to blending artifacts in local regions where pixels share identical NIR values but differ in RGB colors (\textit{e.g.}, the eyes of the owl), resulting in visually degraded RGB restorations. Corresponding quantitative results are reported in Tab.~\ref{tab:ablation_real}. The model with all components achieves the best performance across all metrics and obtains the best scores in human evaluations.

\begin{table}[t]
\centering
\setlength{\abovecaptionskip}{2mm}
\caption{Real-world quantitative ablation results. Best and second best results are annotated with bold and underline.
}
\label{tab:ablation_real}
\setlength{\tabcolsep}{2mm}
\resizebox{0.85\linewidth}{!}{%
\begin{tabular}{lcccc}
\toprule
Models & PI $\downarrow$ & MUSIQ $\uparrow$ & MANIQA $\uparrow$ & HSE $\uparrow$ \\ \midrule
w/o NIR-P.E. & \underline{5.531} & 49.17 & \underline{0.2734} & 3.150 \\
w/o C.C. MLP & 5.586 & \underline{55.73} & 0.2406 & \underline{3.175} \\
Ours & \textbf{5.415} & \textbf{56.51} & \textbf{0.2752} & \textbf{3.450} \\
\bottomrule
\end{tabular}
}
\vspace{-2mm}
\end{table}

\begin{table}[t]
\centering
\setlength{\abovecaptionskip}{2mm}
\caption{Real-world comparisons with ``2D Fusion + NeRF''. 
}
\label{tab:comparison_real_twostage}
\setlength{\tabcolsep}{2mm}
\resizebox{0.85\linewidth}{!}{%
\begin{tabular}{lcccc}
\toprule
Models & PI $\downarrow$ & MUSIQ $\uparrow$ & MANIQA $\uparrow$ & HSE $\uparrow$ \\ \midrule
NVEU + NeRF & \underline{7.746} & \underline{33.74} & 0.2571 & 3.050 \\
SANet + NeRF & 9.571 & \underline{28.27} & 0.2655 & \underline{3.075} \\
NAID + NeRF & 9.359 & 24.96 & \underline{0.2694} & 2.950 \\
Ours & \textbf{5.415} & \textbf{56.51} & \textbf{0.2752} & \textbf{3.450} \\
\bottomrule
\end{tabular}
}
\vspace{-2mm}
\end{table}

\noindent \textbf{Comparison.}
The comparison results on real-world captures are shown in Fig.~\ref{fig:real_comparison} and Tab.~\ref{tab:comparison_real}. Existing methods exhibit limited generalization capability in real-world settings, often leading to texture loss and color distortions. In general, the proposed method achieves superior results, preserving both structural integrity and color fidelity across all scenes.
Additional qualitative comparisons, including those with the “2D Fusion + NeRF” baseline, are provided in the supplementary. The proposed model achieves more accurate RGB recovery, effectively restoring both appearance and geometric details. Quantitative comparison results are presented in Tab.~\ref{tab:comparison_real_twostage}, where the proposed method achieves the best performance.
The proposed method is further evaluated in the RAW domain to demonstrate its generality. 
Specifically, the 12-bit RAW Bayer data is linearly normalized by $4095$, without applying any white-balance or color-correction operations. 
Comparisons are conducted with RawNeRF~\cite{mildenhall2022nerf}, NVEU~\cite{niu2023nir}, and NAID~\cite{xu2024nir}. For NVEU and NAID, the two green channels are averaged, and the resulting RGB values are stacked into a three-channel image. 
Qualitative and quantitative results are reported in Fig.~\ref{fig:rawnerf} and Tab.~\ref{tab:raw}. The results indicate that: (1) the proposed method generalizes effectively to the RAW domain, and (2) it generally achieves superior performance compared to RawNeRF, NVEU, and NAID. White balance postprocessing is not applied for RAW data visualization in Fig.~\ref{fig:rawnerf} to provide a direct and unbiased visualization.

\subsection{Discussions and Limitations}

The model is limited to static scenes, which constrains its applicability in dynamic environments. 
Also, when the RGB and NIR are completely unaligned (\textit{e.g.}, using one RGB camera and another NIR camera for free capture), accurate camera poses for RGB observation are hard to obtain. We leave these for future work.

\begin{table}[t]
\centering
\setlength{\abovecaptionskip}{2mm}
\caption{Real-world quantitative comparison results.
}
\label{tab:comparison_real}
\setlength{\tabcolsep}{2mm}
\resizebox{0.8\linewidth}{!}{%
\begin{tabular}{lcccc}
\toprule
Models & PI $\downarrow$ & MUSIQ $\uparrow$ & MANIQA $\uparrow$ & HSE $\uparrow$ \\ \midrule
Restormer & 8.639 & 16.08 & 0.2333 & 2.550 \\
ScaleMap & 5.776 & 29.61 & 0.1706 & 2.875 \\
NVEU & \textbf{4.817} & 36.01 & 0.2639 & 3.025 \\
SANet & 5.843 & \underline{35.03} & 0.1993 & \underline{3.075} \\
NAID & 9.408 & 19.98 & \underline{0.2732} & 2.925 \\
LLNeRF & 9.721 & 13.23 & 0.2618 & 2.625 \\
Ours & \underline{5.415} & \textbf{56.51} & \textbf{0.2752} & \textbf{3.450} \\
\bottomrule
\end{tabular}
}
\vspace{-2mm}
\end{table}

\begin{table}[t]
\centering
\setlength{\abovecaptionskip}{2mm}
\caption{Quantitative comparison results in RAW space. 
Best and second best results are annotated with bold and underline.
}
\label{tab:raw}
\setlength{\tabcolsep}{2mm}
\resizebox{0.8\linewidth}{!}{%
\begin{tabular}{lccccc}
\toprule
Models & PI $\downarrow$ & MUSIQ $\uparrow$ & MANIQA $\uparrow$ & HSE $\uparrow$\\
\midrule
NVEU & \textbf{4.821} & 46.71 & \underline{0.2482} & 2.850 \\
NAID & 9.924 & 34.76 & 0.2240 & 2.625 \\
RawNeRF & 6.783 & \underline{46.78} & 0.2251 & \underline{3.150} \\
Ours & \underline{5.874} & \textbf{55.70} & \textbf{0.2527} & \textbf{3.775} \\
\bottomrule
\end{tabular}
}
\vspace{-2mm}
\end{table}

\section{Conclusion}
\label{sec:conclusion}

This paper introduces a new 3D-aware model for RGB–NIR dark imaging. Built upon the volume-rendering framework, a novel implicit neural fusion architecture is carefully designed with several effective components. Both synthetic and real-world datasets are provided to demonstrate the superiority and robustness of the proposed model across various scenarios. Without supervision from clean RGB data, the method achieves performance exceeding state-of-the-art approaches, generalizing across different noise levels without using the clean RGB for supervision.

\section*{Acknowledgments}
This work was supported in part by JSPS KAKENHI Grant Number 24KK0209, the Forest Digital Twin Project under the Partnership Agreement for Social Value Creation between UTokyo and SMBC Group, the Hokkaido Sarabetsu Village "Endowed Chair for Field Phenomics" projects in Japan, and the Advanced AI Talent Development to Lead the Next-Generation AI for Intelligent Society (BOOST NAIS) of The University of Tokyo.

\bibliographystyle{ACM-Reference-Format}
\bibliography{sample-base}

\end{document}